\documentclass[letterpaper, 10 pt, conference]{ieeeconf}

\IEEEoverridecommandlockouts                              

\usepackage[english]{babel}
\usepackage[mathletters]{ucs}
\usepackage[utf8]{inputenc}
\usepackage{amssymb,amsmath}
\usepackage[export]{adjustbox}

\usepackage{latexsym}
\usepackage{color} 
\usepackage{indentfirst}
\usepackage{graphicx}       
\usepackage{fancyhdr}
\usepackage{longtable}
\usepackage{pifont}
\usepackage{makeidx}
\usepackage{lastpage}
\usepackage{multirow}
\usepackage{dcolumn} 
\usepackage{epstopdf}
\usepackage{url}
\usepackage{listings}
\usepackage{relsize}
\usepackage{pdfpages}
\usepackage{tabularx}
\usepackage{booktabs}  
\usepackage{siunitx}
\usepackage{float}
\usepackage{textcomp}
\usepackage{hyperref}
\usepackage{afterpage}
\usepackage{tabularx}
\usepackage{bm}  
\usepackage{everypage}

\usepackage[nolist,nohyperlinks]{acronym}
\usepackage{interval}

\usepackage{todonotes}
\usepackage{svg}
\usepackage{algorithmic} 
\usepackage{float}
\usepackage{hyperref}
\usepackage[english]{babel}
\usepackage{csquotes}
\usepackage{subcaption}
\usepackage[T1]{fontenc}

\usepackage{standalone}
\usepackage{pgfplots}
\pgfplotsset{compat=1.16}

\renewcommand\vec{\bm}

\newcommand{\norm}[1]{\left\lVert#1\right\rVert}
\newcommand{\abs}[1]{\left\lvert#1\right\rvert}

\DeclareSIUnit{\pixel}{px}
\DeclareSIUnit{\fps}{FPS}
\DeclareSIUnit{\fish}{fish}

\definecolor{darkgreen}{rgb}{0.0,0.7,0.0}

\acrodef{UAV}{Unmanned Aerial Vehicle}
\acrodef{SWaP}{Size Weight and Power}
\acrodef{CNN}{Convolutional Neural Network}
\acrodef{SGD}{Stochastic Gradient Descent}
\acrodef{UWB}{Stochastic Gradient Descent}

\usepackage{scalerel}

\usetikzlibrary{svg.path}
\definecolor{orcidlogocol}{HTML}{A6CE39}
\tikzset{
  orcidlogo/.pic={
    \fill[orcidlogocol] svg{M256,128c0,70.7-57.3,128-128,128C57.3,256,0,198.7,0,128C0,57.3,57.3,0,128,0C198.7,0,256,57.3,256,128z};
    \fill[white] svg{M86.3,186.2H70.9V79.1h15.4v48.4V186.2z}
    svg{M108.9,79.1h41.6c39.6,0,57,28.3,57,53.6c0,27.5-21.5,53.6-56.8,53.6h-41.8V79.1z M124.3,172.4h24.5c34.9,0,42.9-26.5,42.9-39.7c0-21.5-13.7-39.7-43.7-39.7h-23.7V172.4z}
    svg{M88.7,56.8c0,5.5-4.5,10.1-10.1,10.1c-5.6,0-10.1-4.6-10.1-10.1c0-5.6,4.5-10.1,10.1-10.1C84.2,46.7,88.7,51.3,88.7,56.8z};
  }
}

\newcommand\orcidicon[1]{\href{https://orcid.org/#1}{\mbox{\scalerel*{
        \begin{tikzpicture}[yscale=-1,transform shape]
          \pic{orcidlogo};
        \end{tikzpicture}
}{|}}}}

\let\originalleft\left
\let\originalright\right
\renewcommand{\left}{\mathopen{}\mathclose\bgroup\originalleft}
\renewcommand{\right}{\aftergroup\egroup\originalright}

\newcommand\copyrighttext{%
  \footnotesize \textcopyright~2024 IEEE. Personal use of this material is permitted. Permission from IEEE must be obtained for all other uses, in any current or future media, including reprinting/republishing this material for advertising or promotional purposes, creating new collective works, for resale or redistribution to servers or lists, or reuse of any copyrighted component of this work in other works.}

\newcommand\copyrightnotice{%
\begin{tikzpicture}[remember picture,overlay]
\node[anchor=south,yshift=10pt] at (current page.south) {\fbox{\parbox{\dimexpr\textwidth-\fboxsep-\fboxrule\relax}{\copyrighttext}}};
\end{tikzpicture}%
}

\newcommand\publishedtext{%
  \footnotesize \textcopyright~2024 IEEE International Conference on Robotics and Automation. PREPRINT VERSION - DO NOT DISTRIBUTE. Published version DOI: \href{https://doi.org/10.1109/ICRA57147.2024.10610100}{10.1109/ICRA57147.2024.10610100.}}

\newcommand\publishednotice{%
\begin{tikzpicture}[remember picture,overlay]
\node[anchor=north,yshift=-10pt] at (current page.north) {\publishedtext};
\end{tikzpicture}%
}

\begin{document}

\AddEverypageHook
{
  \publishednotice
  \copyrightnotice
}

\title{Bio-inspired visual relative localization for large swarms of UAVs}

\author{Martin Křížek$^{1,3\orcidicon{0000-0002-2668-4477}}$, Matouš Vrba$^{1\orcidicon{0000-0002-4823-8291}}$, Antonella Barišić Kulaš$^{2\orcidicon{0000-0002-4532-6915}}$, Stjepan Bogdan$^{2\orcidicon{0000-0003-2636-3216}}$ and Martin Saska$^{1\orcidicon{0000-0001-7106-3816}}$%
\thanks{$^{1}$Authors are with the Faculty of Electrical Engineering,
Czech Technical University in Prague, Technická 2, Prague 6,
\url{krizema3@fel.cvut.cz.}}%
\thanks{$^{2}$Authors are with the Laboratory for Robotics and Intelligent Control Systems (LARICS), Faculty of Electrical Engineering  and Computing, University of Zagreb, 10000 Zagreb, Croatia}%
\thanks{$^{3}$This paper is based on the master thesis of the first author https://dspace.cvut.cz/handle/10467/109452.}%
\thanks{This work was funded by CTU grant no SGS23/177/OHK3/3T/13, by the Czech Science Foundation (GAČR) under research project no. 23-07517S, and by the European Union under the project Robotics and advanced industrial production (reg. no. CZ.02.01.01/00/22\_008/0004590). This work was partially funded by the European Union’s Horizon Europe research program Widening participation and spreading excellence, through project Strengthening Research and Innovation Excellence in Autonomous Aerial Systems (AeroSTREAM) - Grant agreement ID: 101071270.}
}


\maketitle

\begin{abstract}
  We propose a new approach to visual perception for relative localization of agents within large-scale swarms of \acp{UAV}.
  Inspired by biological perception utilized by schools of sardines, swarms of bees, and other large groups of animals capable of moving in a decentralized yet coherent manner, our method does not rely on detecting individual neighbors by each agent and estimating their relative position, but rather we propose to regress a neighbor density over distance.
  This allows for a more accurate distance estimation as well as better scalability with respect to the number of neighbors.
  Additionally, a novel swarm control algorithm is proposed to make it compatible with the new relative localization method.
  We provide a thorough evaluation of the presented methods and demonstrate that the regressing approach to distance estimation is more robust to varying relative pose of the targets and that it is suitable to be used as the main source of relative localization for swarm stabilization.
\end{abstract}


\section{Introduction}
A novel marker-less relative localization system for large swarms of \acp{UAV} is presented in this paper.
This approach is inspired by the behavior of large homogeneous animal groups capable of coherent, collision-free and decentralized movement such as fish or insects.
To achieve the impressive large and agile swarms, these animals rely on various sensing modalities (vision, sound, water flow sensors, etc.) but in general they coordinate based on information about the distance and relative bearing of their neighbors \cite{herbert2016understanding}.

To emulate this behavior in robotic systems, various relative localization approaches are currently employed, most of which rely on a detector providing estimates of relative positions of the agent's neighbors.
In this work we focus on monocular visual relative localization because a camera is a light-weight and relatively inexpensive sensor suitable for deployment onboard \ac{SWaP}-constrained autonomous vehicles.
Specifically, because our motivation is developing a large-scale swarm of lightweight and small aerial robots, we focus on marker-less methods that also reduce the payload requirements of the robots.


Marker-less visual relative localization methods typically estimate relative poses of the targets from bounding boxes of their projection onto the image plane provided by a state-of-the-art detection algorithm such as~\cite{tan2020efficientdet}, \cite{centernet} or~\cite{yolov3}.
However, this approach suffers inherent bias in the position estimation because dimensions of the bounding box depend not only on the target's size and distance but also on its relative pose.
Furthermore, detecting individual neighboring team-members and then estimating their accurate relative pose is unnecessarily complex in case of agile swarming and does not scale well to very large numbers of team-members.
Instead, we propose to predict the density of \acp{UAV} in the image over distance from the camera.
As we demonstrate in sec.~\ref{sec:Evaluation}, this enables reducing the number of parameters of the \ac{CNN} and thus faster execution and training while improving accuracy of the predictions.

\subsection{Related work}
Relative localization sensors carried onboard of \acp{UAV} often require markers placed onboard the neighbors to be localized.
Typical examples of such methods are the UVDAR system~ \cite{walter2023difec} that relies on ultraviolet LED markers and UV-sensitive cameras for detection, or the AprilTag system~\cite{apriltag3} and similar methods that detect a black-and-white marker and can estimate its pose relative to the camera given its physical dimensions are known.


Marker-less detection and relative localization is in general a harder problem and marker-less methods are more diverse regarding sensor modality with various sensors compromising between accuracy, robustness, price and \ac{SWaP} restrictions.
There are works relying on a LiDAR sensor~\cite{vrba2023fod},
\ac{UWB} ranging with multilateration~\cite{guo2019ultra},
depth-images from a stereo camera~\cite{vrba_ral2019} and other modalities, but in this work, we focus on light-weight monocular RGB cameras.

Typical vision-based approaches to marker-less relative localization rely on a \ac{CNN}-based object detector estimating bounding boxes of projections of the detected targets to the image plane such as the work presented in~\cite{vrba_ral2020}.
State-of-the-art \ac{CNN} architectures for object detection include approaches focused on maximal detection precision such as YOLO~\cite{yolov3} or CenterNet~\cite{centernet}, or approaches focused on fast execution with limited computational resources like the EfficientDet~\cite{tan2020efficientdet}.
In \cite{pavliv2021tracking}, output of the detector is used to initialize a second \ac{CNN} which predicts positions of selected keypoints in the image that are then used for 6-DoF estimation of the target's pose.
The authors report accuracy up to several centimeters in a small-scale laboratory environment with three \acp{UAV}.
A similar approach is presented in~\cite{bertoni2019monoloco} which proposes a \ac{CNN} regressor to estimate distance of people in an image based on 2-D positions of their joints.
There are also methods for 3-D detection such as~\cite{Brazil_2019_ICCV} where a 3-D bounding box is estimated from RGB data.
However, detecting and estimating the relative pose of individual objects does not scale well to large numbers of targets and are overly complex for swarm stabilization.

As mentioned above, our proposed approach is inspired by biological systems.
It is assumed that e.g. a sardine does not estimate the relative position of every other sardine within its field of view (\cite{sardines2016} reports a mean density of Pacific sardines within a school of \SI{81}{\fish \per \cubic \metre} and smallest school sizes of 5000 fish).
Our method aiming to design a system with similar properties is then relevant to crowd-counting \acp{CNN} such as~\cite{zhang2018crowd, gao2020cnn}.
A typical output of such methods is a 2-D density image of the target objects (typically human heads) within an input RGB image~\cite{sindagi2018survey}.
Tackling accurate density estimation for large as well as small groups, \cite{zhang2018crowd} proposes a relative-count training loss in addition to the commonly employed density map loss.
In the proposed method, a similar loss function is utilized to ensure that even close objects with lower density are accurately detected which is crucial for mutual collision avoidance within a swarm.
In contrast to most crowd-counting methods, we do not estimate a 2-D map of object density in an image, but instead a density over distance from the camera, which is crucial for robust and agile swarming in large UAV groups. To our best knowledge, it is an entirely new approach towards reliable stabilization of compact swarms.

\subsection{Contributions}
Instead of detecting and localizing individual objects in an image, we propose to directly regress their distribution over distance, which is robust to the number of objects being localized.
We demonstrate a \ac{CNN} architecture implementing this approach that outperforms a relative localization approach utilizing bounding boxes from a state-of-the-art detection \ac{CNN} while having less parameters and being faster.
Comparison with an ideal object detector shows that the proposed method outperforms even the best classical approaches detecting the neighbors in the images individually.
Finally, we show that the proposed approach is suitable as a sensing system for swarm stabilization and collision avoidance in several simulated experiments.


\subsection{Problem definition}
\label{sec:ProblemDefinition}
We assume an \ac{UAV} equipped with one or multiple monocular cameras, self-localization sensors, and an onboard computer running the necessary software for autonomous flight.
Let us call it the \textit{focal \ac{UAV}}.
This focal \ac{UAV} observes other \acp{UAV} in the cameras' field of view and uses the output of a visual relative localization algorithm in feedback to control its position relative to the observed neighbors.
The relative localization must be sufficiently accurate and scalable to enable robust, collision-free swarming and it must be efficient enough to process data with a high rate onboard an \ac{UAV} with limited computational capabilities.

\section{Density over distance estimation}
\label{sec:NetworkOutput}
The proposed network architecture is a \ac{CNN} that takes an RGB image represented as a $w_{\text{in}} \times h_{\text{in}} \times 3$ tensor of pixel values as the input.
This image is divided into a regular $w_{\text{out}} \times h_{\text{out}}$ grid.
The output of the network estimates a discrete distribution $l_{\text{raw}}[d]$ of \acp{UAV} over distance for each cell in the grid.
We define $l_{\text{raw}}[d]$ as
\begin{align}
    l_{\text{raw}}[d] = \abs{ \left\{ \vec{p} \mid \vec{p} \in \mathcal{M},~ \norm{\vec{p}} \in [d,d + {}_\Delta d]~\right\} }, \label{eq:cnnOut1}
\end{align}
where $d \in \left\{ 0,~ {}_\Delta d,~ 2{}_\Delta d,~ 3{}_\Delta d~ \dots \right\}$ is a distance from the camera, ${}_\Delta d$ is a distance discretization step, $\mathcal{M}$ is a set of positions of \acp{UAV} in the corresponding image grid cell, and $\vec{p}$ is a position of an \ac{UAV} (expressed in the camera's optical frame).

\begin{figure}
  \centering
  \begin{subfigure}[]{0.73\linewidth}
    \centering
    \includegraphics[width=1.0\linewidth]{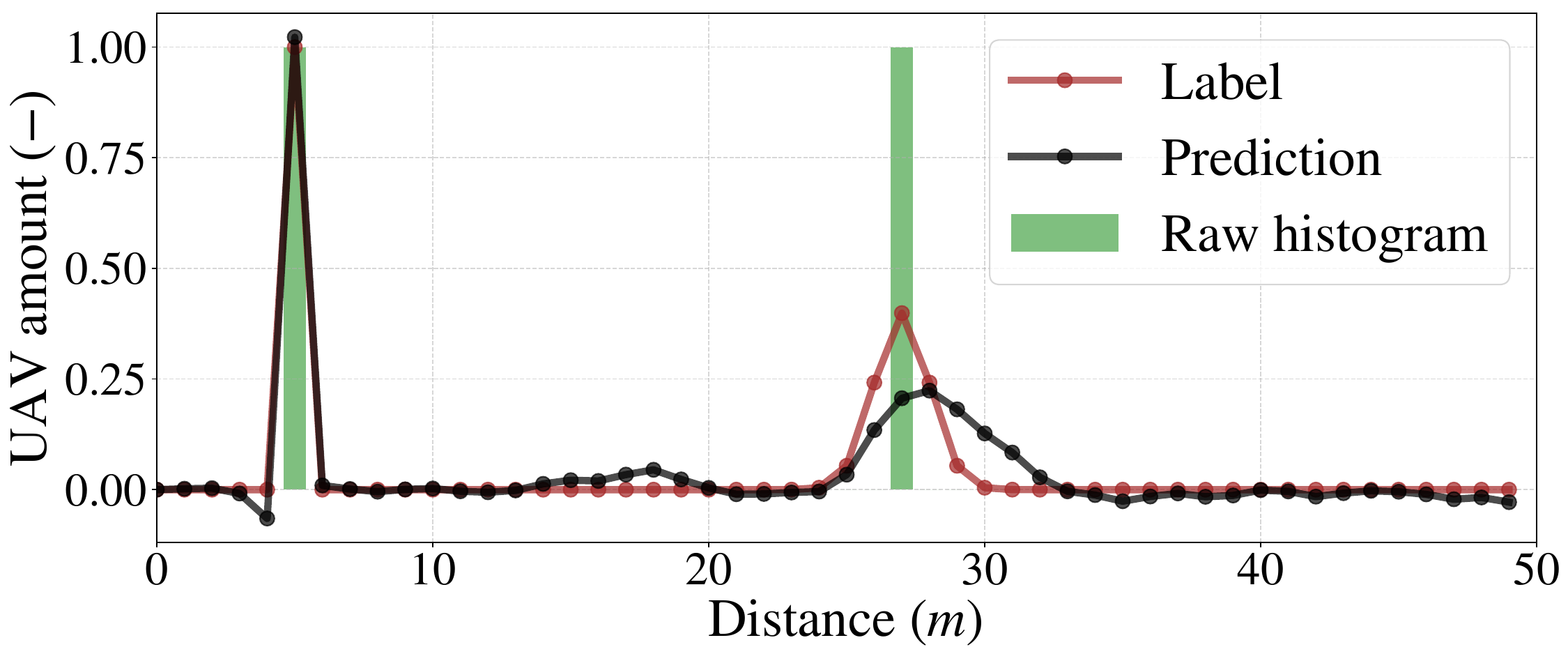}
  \end{subfigure}
  \hfill
  \begin{subfigure}[]{0.24\linewidth}
    \centering
    \includegraphics[width=1.0\linewidth]{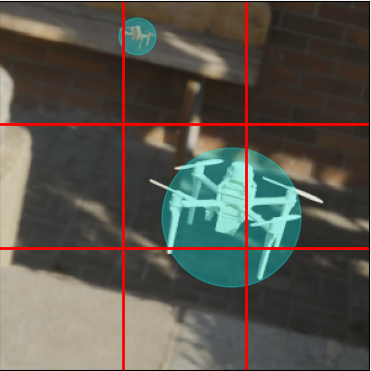}
    \vspace{0.01cm}
  \end{subfigure}  
  \begin{subfigure}[]{0.7\linewidth}
    \centering
    \includegraphics[width=1.0\linewidth]{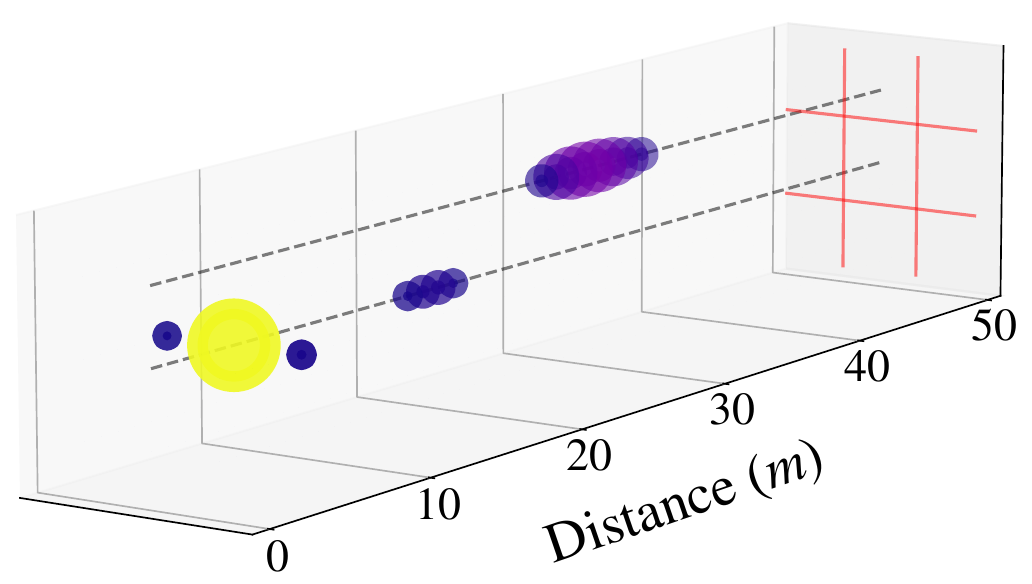}
    \vspace{0.01cm}
  \end{subfigure}
  \caption{%
\textbf{Top:}
An illustration of the Gaussian smoothing and the \ac{CNN} prediction of the distribution of \acp{UAV} in the image over distance.
The input image is shown on the right.
\textbf{Bottom:}
The output of the \ac{CNN} with $w_{\text{out}} = h_{\text{out}} = 3$ for the input image.
The predicted density for a given grid cell and distance is marked with a circle of the corresponding size and color from blue to yellow.
}\label{fig:lab_to_gauss_166}
\vspace{-1em}
\end{figure}

The function $l_{\text{raw}}[d]$ can be interpreted as a histogram where each bin represents a distance range with a width ${}_\Delta d$ (i.e. for ${}_\Delta d = \SI{1}{\metre}$, the first bin represents the distance range from $\SI{0}{\metre}$ to $\SI{1}{\metre}$ etc.) and the value of $l[d]$ represents the estimated amount of \acp{UAV} in the image within the corresponding range (see Fig.~\ref{fig:lab_to_gauss_166}).
For practical reasons, the number of discretization bins is limited to $n_{\text{bin}}$ and \acp{UAV} beyond $d_{\text{max}} = {}_\Delta d \left( n_{\text{bin}} - 1 \right)$ are considered to belong to the last bin:
\begin{equation}
    l_{\text{raw}}[d_{\text{max}}] = \abs{ \left\{ \vec{p} \mid \vec{p} \in \mathcal{M},~ \norm{\vec{p}} > d_{\text{max}} \right\} }. \label{eq:cnnOut2}
\end{equation}

When training the \ac{CNN}, we do not use the raw histogram $l_{\text{raw}}[d]$ directly for the ground-truth labels, but rather a partially smoothed real-valued function $l_{\text{gt}}[d]$ obtained as
\begin{equation}
    l_{\text{gt}}[d] = \left( l_{\text{raw}} * g \right)[d], \label{eq:cnnOut3}
\end{equation}
\begin{equation}
    g[d] = \begin{cases}
        \delta[d], & \text{for the closest $k$ bins}, \\
        G_{\sigma}[d], & \text{otherwise},
  \end{cases}
\end{equation}
where $G_{\sigma}[d]$ denotes a Gaussian kernel with a standard deviation $\sigma$, $\delta[d]$ is the identity kernel, and $\left( f * g \right)[x]$ denotes discrete convolution of functions $f[x]$ and $g[x]$ w.r.t. $x$.
As discussed in sec.~\ref{sec:Data} and sec.~\ref{sec:Evaluation}, this significantly improves training convergence and overall performance of the \ac{CNN}.
Formally, the domain and range of $l_{\text{gt}}[d]$ are
\begin{align}
    l_{\text{gt}} &: \mathcal{D} \rightarrow \mathbb{R}, \\
    \mathcal{D} &= \left\{ 0,~ {}_\Delta d,~ 2{}_\Delta d,~ \dots,~ d_{\text{max}} \right\},~ {}_\Delta d \in \mathbb{R},~ d_{\text{max}} \in \mathbb{R}. \nonumber
\end{align}
The network's output $l_{\text{o}}[d]$ is then trained to estimate $l_{\text{gt}}[d]$.

For mathematical conciseness, let us interpret the network's output $l_{\text{o}}[d]$ as an $n_{\text{bin}}$-dimensional vector $\vec{l}_{\text{o}} \in \mathbb{R}^{n_{\text{bin}}}$ so that the $i$-th element of $\vec{l}_{\text{o}}$ is equal to $l_{\text{o}}[{}_\Delta d \left( i - 1 \right)]$.
We shall denote this as
\begin{align}
    \vec{l}_{\text{o}} = l_{\text{o}}[d].
\end{align}

\subsection{Network}\label{sec:Architecture}
\begin{figure}
  \centering
  \includegraphics[width=1\linewidth]{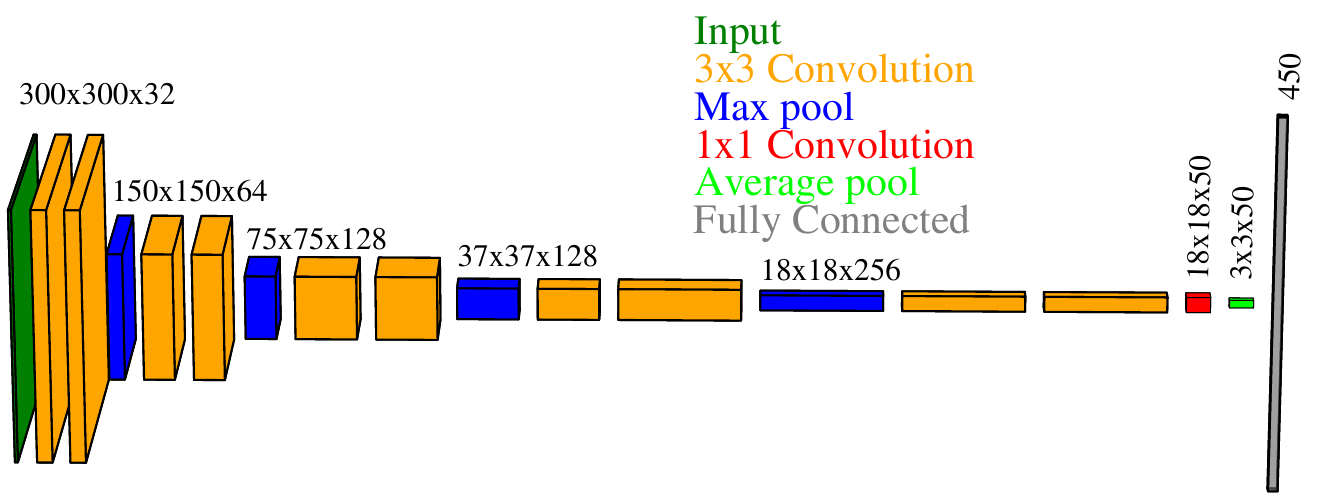}
  \caption{Visualization of the network architecture for $w_{\text{in}} = h_{\text{in}} = 300$, $w_{\text{out}} = h_{\text{out}} = 3$, and $n_{\text{bin}} = 50$.}\label{fig:netarch}
\vspace{-1em}
\end{figure}

The neural network (see Fig.~\ref{fig:netarch}) consists of a feature extraction head followed by a custom tail with the output as specified in the previous section.
For the head, a VGG-like feature extractor structure \cite{simonyan2014very} was employed.

The head is followed by a single convolutional layer with a kernel of size $1 \times 1$ and a filter dimension of $n_{\text{bin}}$.
This $1 \times 1$ convolutional layer similarly as in the related crowd counting approaches \cite{zhang2016single, zhang2018crowd} weights the feature maps according to their importance for every distance bin in the output vector.
Moreover, utilizing the $1 \times 1$ convolution instead of dense layers, which are often used at the top of the network, proved to have better performance with a significantly reduced number of parameters which is crucial for online deployment.

The output of this layer is connected to an average pooling layer that produces $w_{\text{out}} \times h_{\text{out}}$ vectors of dimension $n_{\text{bin}}$ which are then connected via a single fully connected layer (consisting of $w_{\text{out}} \times h_{\text{out}} \times n_{\text{bin}}$ nodes) to the output of the network.
These last layers facilitate regression of the output \ac{UAV} density vectors $\vec{l}_{\text{o}}$ for each grid cell from the features acquired by the feature extractor.

\subsection{Loss function}
\label{sec:LossFunction}
A weighted Euclidean distance between the predicted densities and ground-truth labels was employed as the loss function for training and evaluation of the \ac{CNN}.
Assuming $w_{\text{out}} = h_{\text{out}} = 1$, the error function for a single image is then defined as
\begin{equation}
    \mathcal{L} = \norm{ \left(\vec{l}_{\text{o}} - \vec{l}_{\text{gt}} \right) \circ \vec{w} }, \label{eq:loss}
\end{equation}
where $\vec{l}_{\text{gt}}$ is the corresponding ground-truth density vector, $\vec{w}$ is a vector of (non-learnable) bin weights, and the symbol $\circ$ represents element-wise multiplication between vectors.
For image grids with $w_{\text{out}},~ h_{\text{out}} > 1$, the vectors $\vec{l}_{\text{o}}$, $\vec{l}_{\text{gt}}$, and $\vec{w}$ for all cells are stacked into meta-vectors $\vec{L}_{\text{o}}$, $\vec{L}_{\text{gt}}$, $\vec{W}$ and then the loss function is calculated analogically to eq.~\eqref{eq:loss} using these meta-vectors.

The weight vector $\vec{w}$ scales the errors of each distance bin which allows to target specific distance levels that are of higher interest -- typically those that are close to the camera (and thus the focal \ac{UAV}).
Having reliable detections near the focal \ac{UAV} is crucial for mutual collision avoidance and collision-free swarming.
However, the majority of the \acp{UAV} in our training dataset are far away from the focal drone as discussed in sec.~\ref{sec:Data}.
The weight vector $\vec{w}$ counters this bias in the data and significantly improves the detection success rate of the nearby \acp{UAV} (see sec.~\ref{sec:Evaluation}).

\subsection{Training setup}
\label{sec:Training}
The dataset described in the following section was split into $10000$ training, $1000$ validation, and $5000$ testing samples.
The training data was used for the \ac{SGD} back-propagation with a batch size of 32 samples.
During the training, we gradually decreased the starting learning rate of $0.001$ by $20\%$ if the validation loss has not decreased by at least $0.0001$ in the last 5 epochs.
After 60 epochs, the weights with the lowest loss on the validation data were selected as the best result.
The testing data was used for comparing different variants of the \ac{CNN} model (see sec.~\ref{sec:Evaluation}).
We used the ADAM \cite{kingma2014adam} optimizer with the loss function defined in the previous section for the training. 
The Tensorflow library \cite{abadi2016tensorflow} was employed to implement, train and test our neural network model.

\section{Dataset}\label{sec:Data}
To our best knowledge, a sufficiently sizable, properly labeled dataset with the large numbers of \acp{UAV} required to fully utilize the presented method is not currently publicly available.
Therefore, we opted to generate a realistic synthetic dataset of $16 000$ images using the approach introduced in \cite{barisic2022sim2air}.
The images were generated with 272 HDRI maps for background including natural and urban environments with a wide variety of lighting conditions.
To improve the detector's ability to handle challenging environmental conditions and different camera sensors, we applied different color textures to the \acp{UAV} as described in~\cite{barisic2022sim2air}.

\begin{figure}
    \centering
    \begin{subfigure}{0.38\linewidth}
        \centering
        \includegraphics[width=0.7\linewidth]{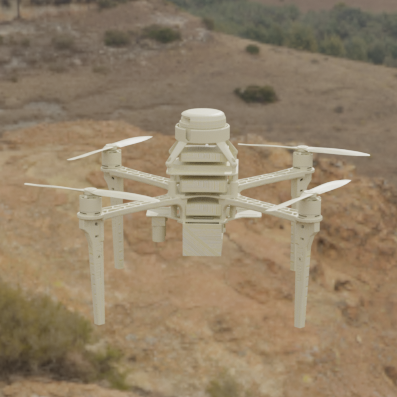}
        \caption{Synthetic image.}
    \end{subfigure}
    ~
    \begin{subfigure}{0.38\linewidth}
        \centering
        \includegraphics[width=0.7\linewidth]{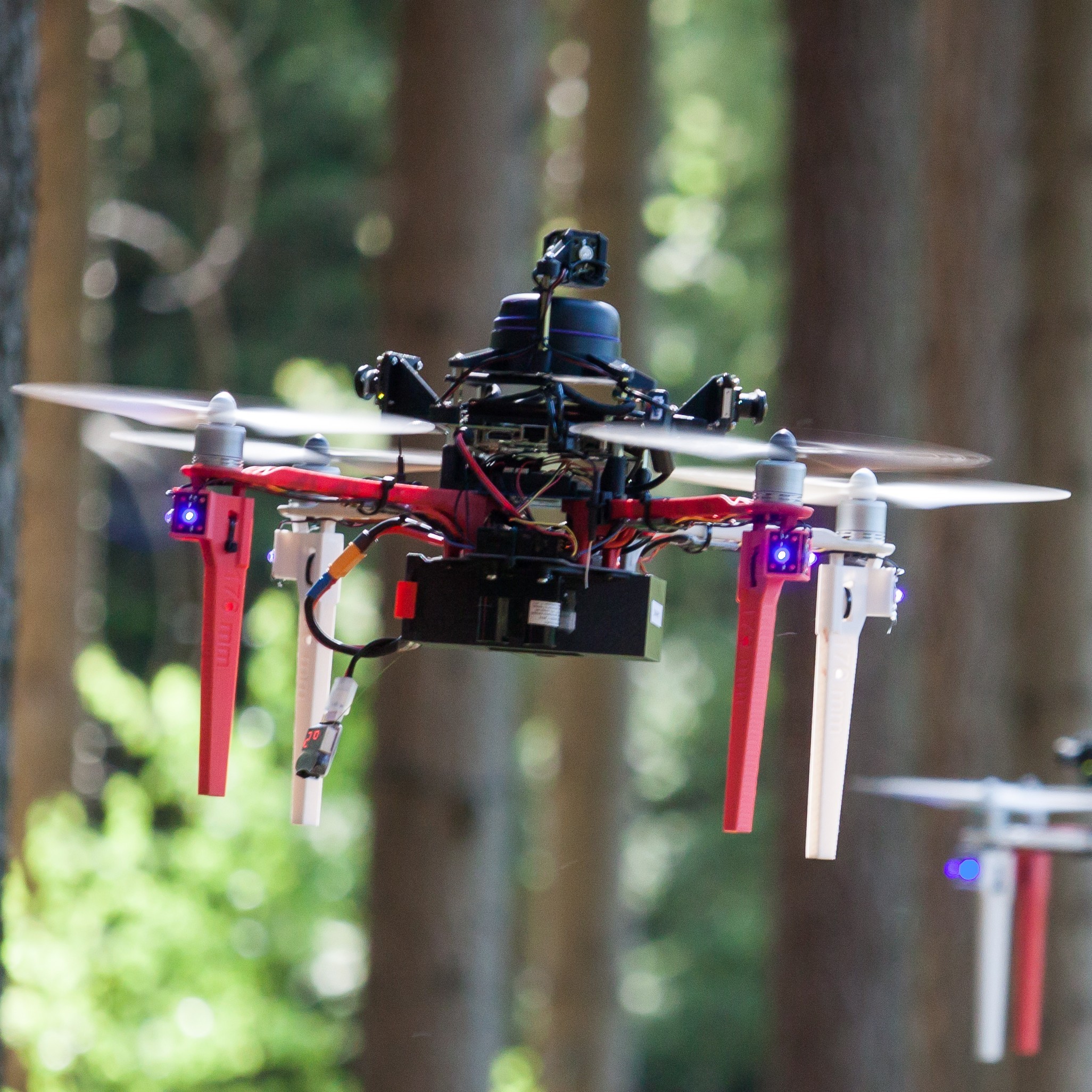}
        \caption{Real-world photo.}
    \end{subfigure}
    \caption{Comparison of synthetic (a) and real (b) images of the MRS F450 platform \cite{MRS:platform} used in this work.}
    \label{fig:synth_real_comparison}
\vspace{-1em}
\end{figure}

The MRS F450 \ac{UAV} platform \cite{MRS:platform} was selected for the dataset as well as the real-world swarming experiments.
A 3D model of the platform was designed in Blender and used to generate the data.
Our work expands upon~\cite{barisic2022sim2air}, which only involved up to 4 \acp{UAV}, by developing a procedural pipeline capable of generating \ac{UAV} groups within the scene and containing a random number of targets.
To avoid introducing too many occlusions, we create only up to one group in close proximity to the camera sensor in the virtual scene, while further away, we create multiple groups positioned randomly.
We opt for random generation to maintain a high level of data variance.
An example image from the dataset and a comparison with a real-world photo of the platform during the experimental deployment can be seen in Fig.~\ref{fig:synth_real_comparison}.

The images used for the dataset crops were originally generated with optical parameters corresponding to the color camera of the Intel RealSense D435, which is carried onboard the MRS F450 platforms.

\begin{figure}
  \centering
  \includegraphics[width=1\linewidth]{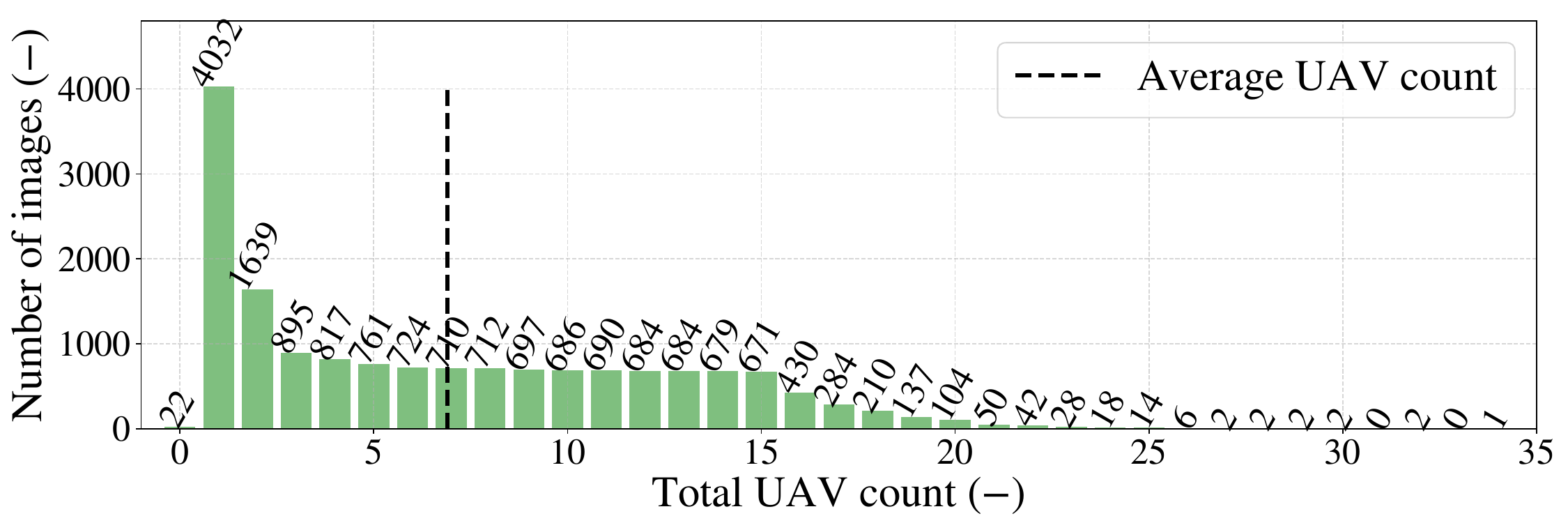}
  \caption{Distribution of images in the dataset based on the total number of \acp{UAV} in the given image.
  The exact counts are displayed above the bars.
  }\label{fig:groupSizeDist}
\vspace{-1.5em}
\end{figure}

Because of a finite resolution of the simulated camera and non-zero dimensions of the \acp{UAV}, images with higher numbers of \acp{UAV} are less available within a $300\times 300$ pixel crop (see Fig.~\ref{fig:groupSizeDist}).
Enforcing the dataset balancing for all numbers of \acp{UAV} would thus degrade its variability, so we only enforce it up to an empirically determined amount of 15 targets as a trade-off between uniformity of the resulting dataset and sufficient data variability.

\begin{figure}
\vspace{-1em}
  \centering
  \includegraphics[width=1\linewidth]{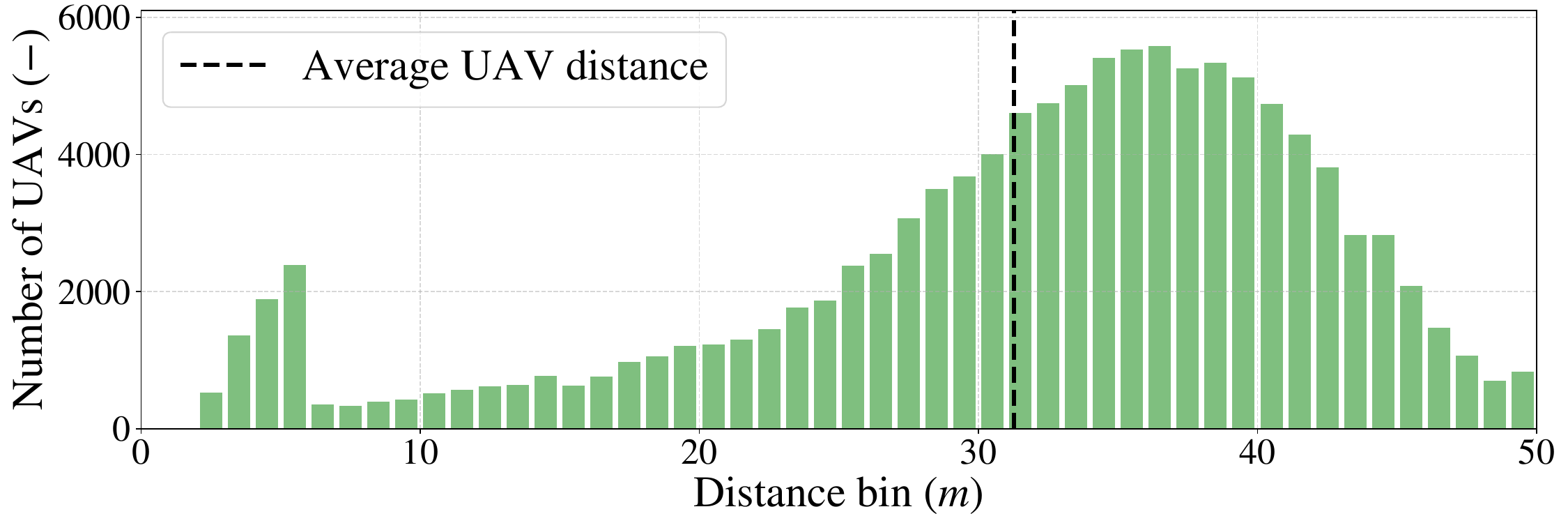}
  \caption{Distribution of \acp{UAV} in the dataset images based on their corresponding distance bin. }\label{fig:binDist}
\end{figure}

Furthermore, the image cropping was also biased to select more crops with \acp{UAV} near the camera (see Fig.~\ref{fig:binDist}) because correctly detecting and estimating the distance of close targets is crucial for safe swarming and collision-avoidance.
This causes images with a single or two \acp{UAV} to be overrepresented in the resulting dataset (see Fig.~\ref{fig:groupSizeDist}) due to the nature of the camera sensor.

\subsection{Labeling}
\label{sec:Labeling}
Ground-truth labels for the images in the dataset were generated using known poses of the \acp{UAV} in the image relative to the camera according to eqs.~(\ref{eq:cnnOut1})-(\ref{eq:cnnOut3}).
As expressed by eq.~(\ref{eq:cnnOut3}), we do not train the \ac{CNN} directly on the ground-truth histogram of \ac{UAV} count over distance, but rather smooth the \ac{UAV} distribution by a Gaussian kernel, which is inspired by related works on crowd counting \cite{zhang2016single, zhang2018crowd}.
The kernel is normalized so that the sum of the distribution does not change and thus the total number of \acp{UAV} in the image is the same as for the non-smoothed label.
Such Gaussian smoothing of the labels reduces the penalization of cases when the \ac{CNN} makes a minor error in distance estimation for some neighbors and shifts them to a nearby bin.
This significantly improves training convergence and the overall performance of the \ac{CNN} (see sec.~\ref{sec:Ablation}).

The dataset was also labeled for standard object detection algorithms by calculating an axis-aligned bounding box of the projection of each \ac{UAV} to the camera image.
The whole dataset including the ground-truth labels is available online\footnote{\href{http://mrs.felk.cvut.cz/perception-for-swarming2023}{\url{mrs.felk.cvut.cz/perception-for-swarming2023}}}.


\subsection{Real-world data}
\label{sec:RealData}
To evaluate capability of the proposed architecture and artificial dataset to generalize to real-world data, we have created a manually annotated dataset of $40$ photos from our real-world experiments with \acp{UAV}.
Because the ground-truth position of the targets in these images is not available, we have employed the nearest-neighbor method for distance estimation described in sec.~\ref{sec:ObjectDetectionComparison}.
The dataset contains between $1$ to $8$ \acp{UAV} per image in various environments (desert, forest, field) and some of the targets have slightly different hardware configuration (and therefore also visual appearance) than used in training.
The evaluation results are described in sec.~\ref{sec:Evaluation}.


  

\section{Evaluation}
\label{sec:Evaluation}
To quantify the performance of the proposed method, we use two metrics: an average absolute per-bin error $\bar{e}[d]$, and a total integral error $\bar{T}$.
Using notation from the previous sections, these errors are defined as
\begin{align}
    e^i[d] &= l^i_{\text{o}}[d] - l^i_{\text{raw}}[d], &&
    \bar{e}[d] = \frac{ \sum_{i \in \mathcal{I}} \abs{ e^i[d] } }{ \sum_{i \in \mathcal{I}} l^i_{\text{raw}}[d] }, \\
    T^i &= \frac{\abs{ \sum_{d\in\mathcal{D}} e^i[d] }}{\sum_{d\in\mathcal{D}} l^i_{\text{raw}}[d]}, &&
    \bar{T} = \sum_{i \in \mathcal{I}} \frac{ T^i }{ \abs{\mathcal{I}} },  \label{eq:errorDefinitions}
\end{align}
where $i$ is a single image from the input dataset $\mathcal{I}$ and the upper index denotes a value for the specific image.

The average absolute per-bin error $\bar{e}[d]$ can have two fundamental sources: misdetections (false positives and false negatives), and incorrect estimation of the distance of detected objects.
These two causes represent different problems but cannot be distinguished solely based on the per-bin error which is why we also measure the total integral error $\bar{T}$.
Together, these measures allow us to estimate which of the above-mentioned causes is the main source of error for a given image.
A lower $\bar{T}$ indicates that most of the per-bin error is caused by inaccuracy in the distance estimation whereas a higher $\bar{T}$ indicates misdetections of objects in the image.
The errors are normalized by the \ac{UAV} density so that they are suitable for comparing the \ac{CNN}'s performance between images with different numbers of \acp{UAV}.

For easier comparison of the different relative localization methods, we define two additional summary metrics
\begin{align}
    \bar{E} = \sum_{d\in \mathcal{D}}  \bar{e}[d],
    && \bar{E}' = \sum_{d=2}^{11} \bar{e}[d],
\end{align}
where $\bar{E}$ is the total average per-bin error and $\bar{E}'$ is the total per-bin error for close distances which we have empirically determined to be crucial for collision-free swarming.

Table~\ref{tab:ablation} presents comparison of the different relative localization methods considered in sections \ref{sec:Testing1D}-\ref{sec:ObjectDetectionComparison} using the previously defined metrics.
We also state the relative number of learnable parameters and the measured average execution time $t_{\text{exec}}$ from 100 inference runs on a GeForce GTX 1050M GPU.
All \acp{CNN} variants in these sections were trained and tested using the dataset discussed in sec.~\ref{sec:Data}.

\subsection{Grid division comparison}
\label{sec:Testing1D}
\begin{figure}
  \centering
  \includegraphics[width=1.0\linewidth]{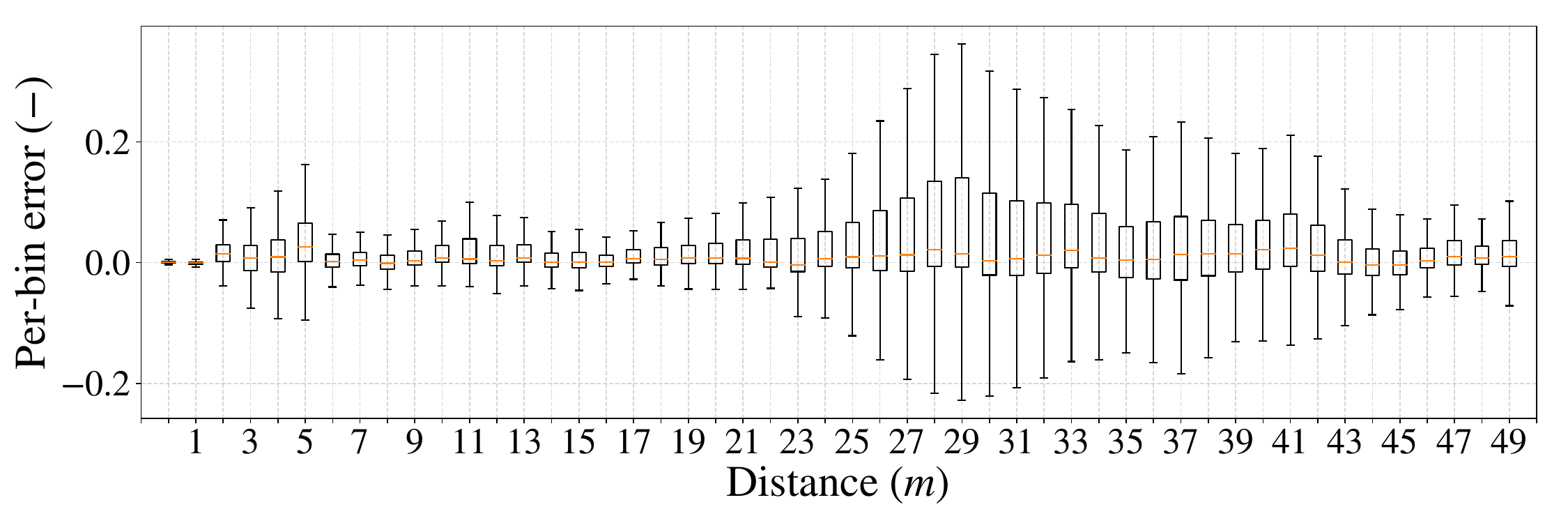}
  \caption{A boxplot graph of the per-bin errors $e^i[d]$ for $w_{\text{out}} = h_{\text{out}} = 1$. }\label{fig:25w_nonscaled_boxplot_paper} 
\vspace{-1em}
\end{figure}

Let us first examine the output of the \ac{CNN} for a case with a single cell in the output grid (meaning that $w_{\text{out}} = h_{\text{out}} = 1$) to evaluate its distance estimation capabilities.
The \ac{CNN} was trained as described in sec.~\ref{sec:Training}.
Output of the \ac{CNN} on the testing dataset was then used to calculate the metrics defined in eq.~\eqref{eq:errorDefinitions} for each data sample.
Fig.~\ref{fig:25w_nonscaled_boxplot_paper} shows the resulting distribution of the per-bin errors $e^i[d]$.
Although the distribution of the error depends on the distribution of the \acp{UAV} in the dataset, it may still be observed that the error is symmetrical and the \ac{CNN} does not significantly bias towards over- or under-estimating the number of objects in a bin.
The total average absolute per-bin error was $\bar{E} = 1.31$ and the total integral error was $\bar{T} = 0.142$, so we conclude that the main source of the per-bin error is not misdetection but rather inaccuracies in distance estimates.

\begin{figure}
  \centering
  \includegraphics[width=1.0\linewidth]{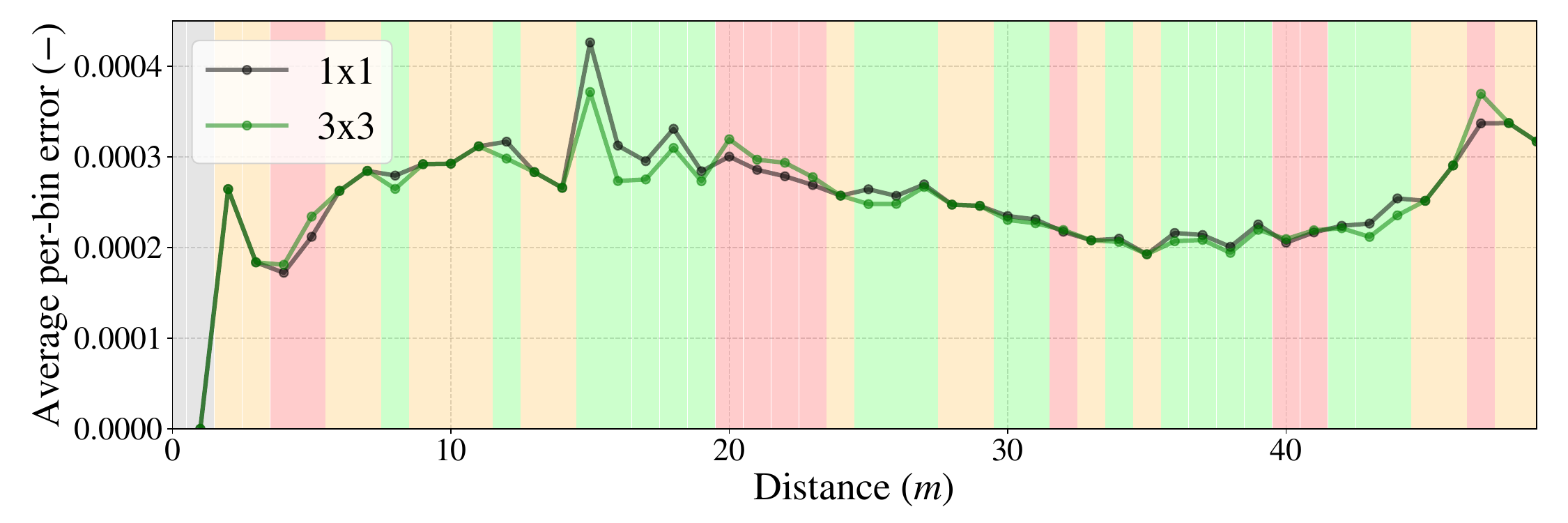}
  \caption{Comparison of the average absolute per-bin error for the $1\times 1$ and $3\times 3$ output grid divisions. (green if $1 \times 1$ is better, red if $3 \times 3$ is better, orange if equal)}\label{fig:1dVS3d} 
  \vspace{-1em}
\end{figure}

Next, we evaluate the proposed \ac{CNN} for a 3D localization scenario with $w_{\text{out}} = h_{\text{out}} = 3$.
As evident from Fig.~\ref{fig:1dVS3d}, the performance is almost identical for the $1\times 1$ and $3\times 3$ division.
The total errors were also similar to the $1\times 1$ version as evident from Table~\ref{tab:ablation}.
The more granular grid division increases the number of learnable parameters of the architecture by approx. 10\%. Therefore, we conclude that estimating the density vectors for multiple cells at once does not negatively impact the performance.

\begin{figure}[] 
  \centering
  \hfill
  \begin{subfigure}[]{0.29\linewidth}
    \centering
    \includegraphics[width=1.0\linewidth]{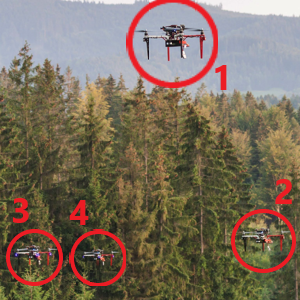}
    \caption{The input image.}
  \end{subfigure}
  \hfill
  \begin{subfigure}[]{0.39\linewidth}
    \centering
    \includegraphics[width=1.0\linewidth]{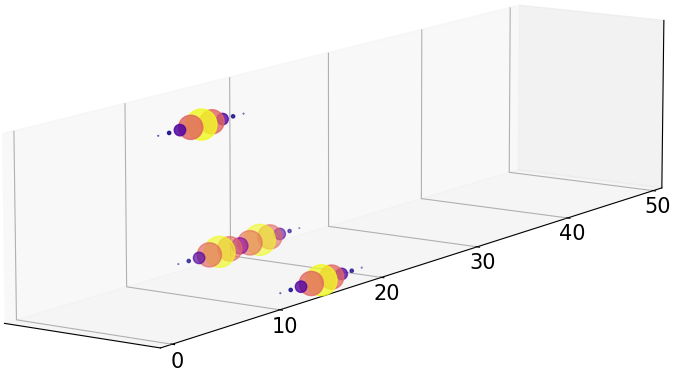}
    \caption{Ground truth.}
  \end{subfigure}
  \hfill

  \hfill
  \begin{subfigure}[]{0.39\linewidth}
    \centering
    \includegraphics[width=1.0\linewidth]{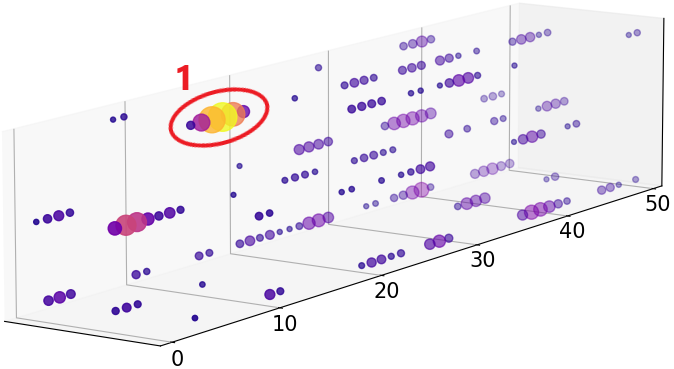}
    \caption{Only synthetic data.}
  \end{subfigure}
  \hfill
  \begin{subfigure}[]{0.39\linewidth}
    \centering
    \includegraphics[width=1.0\linewidth]{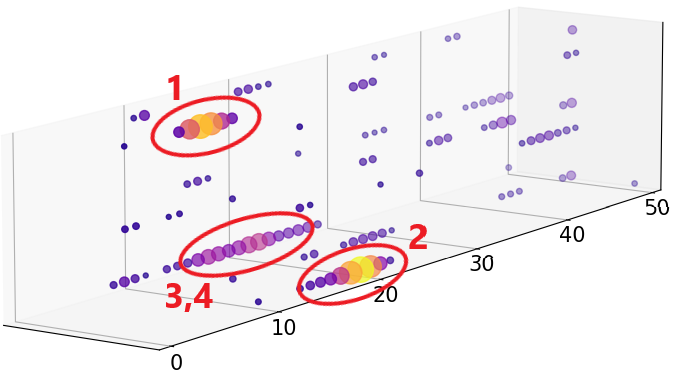}
    \caption{Fine-tuned on real-world data.}
  \end{subfigure}
  \hfill
  \caption{%
  Comparison of the output of the \ac{CNN} on a real-wold photo for different training data.
  The longest axis represents the distance in meters.
  The amount of \acp{UAV} at a given distance and division is represented by color and size of the points.
  Detections corresponding to the ground-truth \ac{UAV} positions are highlighted with red ellipses and enumerated.}\label{fig:real2}
\vspace{-1em}
\end{figure}

\begin{figure}
  \vspace{-1em}
  \centering
  \begin{subfigure}[b]{0.64\linewidth}
    \centering
    \includegraphics[width=1.0\linewidth]{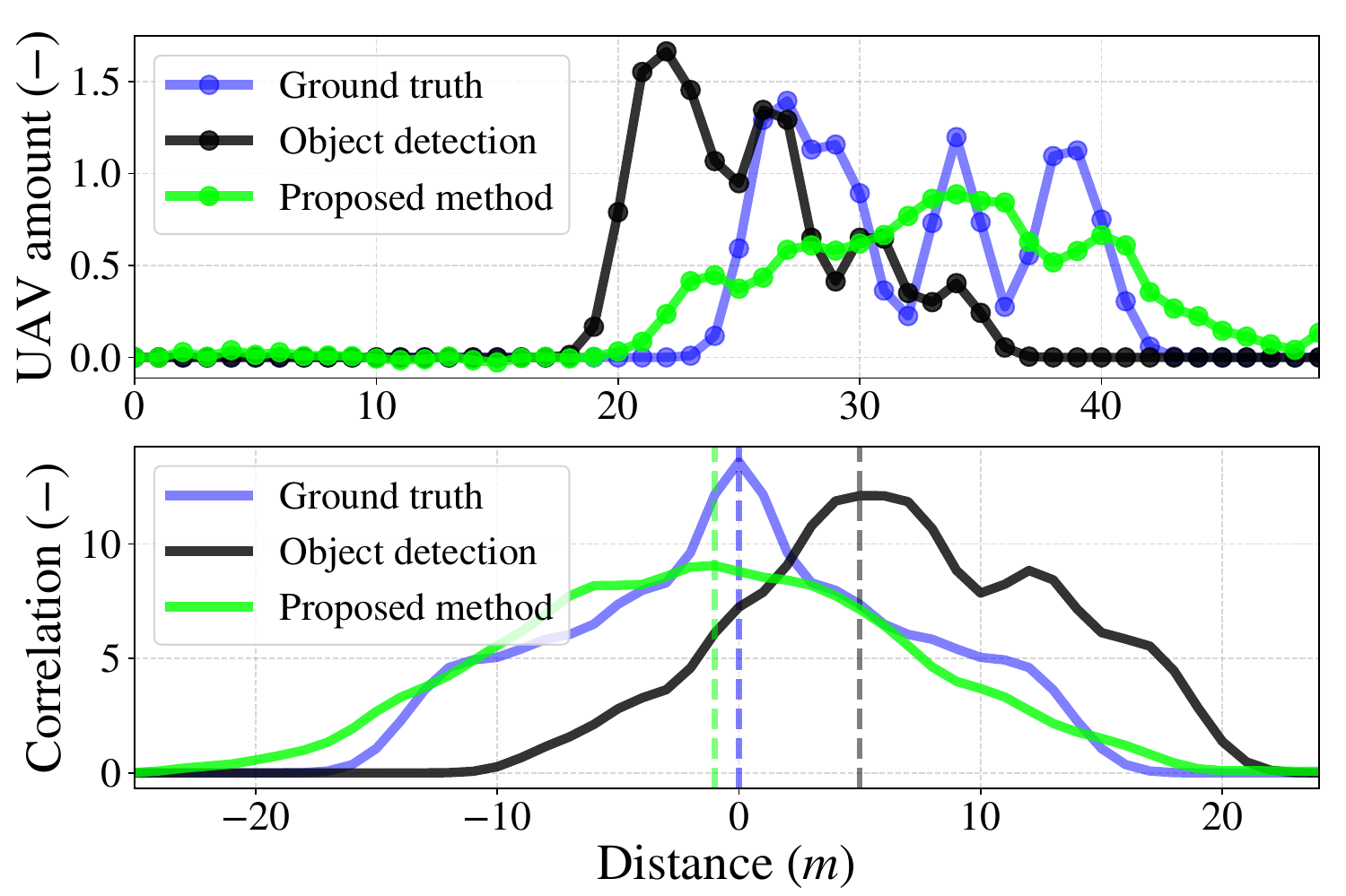}
  \end{subfigure}
  \hfill
  \begin{subfigure}[b]{0.34\linewidth}
    \centering
    \includegraphics[width=1.0\linewidth]{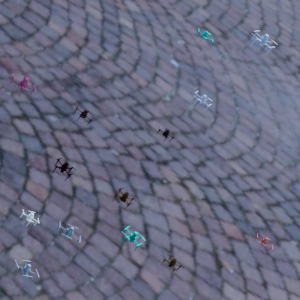}
    \vspace{0.15cm}
  \end{subfigure}  
  \caption{%
An example of a challenging image for bounding box-based distance estimation.
The top graph shows the predicted and ground-truth \ac{UAV} densities over distance.
The bottom graph presents a correlation (as in signal processing) between the predictions and the ground truth.
The highest correlation value is marked with a vertical line.
}\label{fig:tilt1}
  \vspace{-1em}
\end{figure}

\subsection{Ablation study}
\label{sec:Ablation}
An ablation study of the various design choices made for the proposed \ac{CNN} architecture was performed.
The influence of using a $1\times 1$ convolution kernel as the second-last layer instead of the more conventional choice of a fully connected layer was tested as well as an alternative labeling method to the one described in sec.~\ref{sec:Labeling}.
Namely, we trained the \ac{CNN} with the raw labels $l_{\text{raw}}[d]$ (see eq.~\eqref{eq:cnnOut1}), partially smoothed labels $l_{\text{gt}}[d]$ (eq.~\eqref{eq:cnnOut2}) and fully smoothed labels
\begin{equation}
    l_{\text{s}}[d] = \left( l * G_{\sigma} \right)[d].
\end{equation}

Although using only the smoothed labels improves the total integral error $\bar{T}$, indicating a lower misdetection rate, the distance estimation accuracy is reduced -- esp. in the interval close to the observer as evident by the increased error $\bar{E}'$ (see Table~\ref{tab:ablation}).
This may not be concerning for some applications, but for the swarm control considered in this work, we choose the combined smoothing as the best labeling method.

\subsection{Object detection comparison}
\label{sec:ObjectDetectionComparison}
To show that the proposed approach is not only viable but also performs better than an state-of-the-art detection \ac{CNN}-based algorithms such as \cite{tan2020efficientdet, centernet, yolov3}, we compare our method to a YOLOv4 Tiny detector with an added YOLO layer.
Furthermore, to accommodate for possible future advances in object detection \ac{CNN} architectures, we also compare the results with a hypothetical ideal object detector which we assume to output predicted bounding boxes of individual \acp{UAV} in the image with absolute precision.

For each distance bin, the average width and height of the ground-truth bounding boxes in the training dataset are calculated.
Distance of \acp{UAV} is then estimated from the detector's output based on dimensions of the predicted bounding boxes by comparing it with the average bounding box width and height for each distance bin.
The resulting estimated distance is then the nearest-neighbor bin.
The width and height of the bounding box can differ for a single \ac{UAV} at a given distance if the relative pose of the \ac{UAV} to the camera changes (e.g. a different pitch or roll rotation).
This causes an inherent error when estimating the distance of a non-spherical target solely from its bounding box even for an absolutely accurate bounding box prediction.

The hypothetical ideal detector performs the best for close ranges but our method has a better total accuracy distance because it learns to regress the distance directly from the visual features and thus does not suffer from this inherent error.
An illustrative example is shown in Fig.~\ref{fig:tilt1}, where the observing (focal) \ac{UAV} is tilted.
It may be seen in the correlation graph in the Figure that our method has a lower bias in the distance estimation as the correlation maximum is closer to zero.
The hypothetical detector has a zero total integral error $\bar{T}$, but this result has low significance because we did not consider misdetections by the ideal detector.

Our method outperforms the state-of-the-art detector using all metrics because not all of the \acp{UAV} are successfully detected by the EfficientDet \ac{CNN} and the bounding boxes are less accurate which negatively influences the distance estimate.
Furthermore, our method has a lower number of learnable parameters and a faster execution time.

\begin{table}
    \centering
    \begin{tabularx}{0.48\textwidth}{ X l l l l l }
        \textbf{\ac{CNN} variant} & $\bar{T}$ & $\bar{E}$ & $\bar{E}'$ & params & $t_{\text{exec}}$ \\\hline
        Ours ($3\times 3$) & 0.174 & $1.30$ & $1.27$ & $100\%$ & \SI{18}{\milli \second}  \\
        Ours ($1\times 1$) & $0.142$ & $1.31$ & $1.25$ & 91\% & \SI{7}{\milli \second} \\
        F.C. layer ($3\times 3$) & $0.261$ & $1.60$ & $1.71$ & $285\%$ & \SI{8}{\milli \second} \\
        raw labels ($3\times 3$) & $0.410$ & $1.53$ & $1.42$ & $100\%$ & \SI{18}{\milli \second} \\
        smooth labels ($3\times 3$) & $\mathbf{0.119}$ & $1.34$ & $1.55$ & $100\%$ & \SI{18}{\milli \second} \\
        SotA detector & $0.02$ & $\mathbf{1.31}$ & $\mathbf{0.93}$ & N/A & N/A \\
        ideal detector  & $0$ (N/A) & $1.35$ & $0.99$ & N/A & N/A 
    \end{tabularx}
    \caption{%
Comparison of all considered variations of the relative localization approach using various metrics.
The two best results are highlighted in bold.
}
    \label{tab:ablation}
  \vspace{-1.5em}
\end{table}

\subsection{High density range data}

The following section evaluates the performance of the proposed model on data with high density range (difference between the numbers of \acp{UAV} in the images with the least and the most \acp{UAV}). 
The density range of the dataset described in sec.~\ref{sec:Data} is from 1 \ac{UAV} up to 35 \acp{UAV} per image. 
We generated an additional dataset containing all counts from 1 \ac{UAV} up to 150 \acp{UAV} per image. 
We both trained and tested the proposed model on the high density range dataset to demonstrate the scaling capability of the proposed method. 
We trained and tested the object detector on the additional dataset as well to compare with the results of the proposed method. 
Table~\ref{tab:densityrange} presents the results and comparison of both methods.

\begin{table}
    \centering
    \begin{tabularx}{0.48\textwidth}{ X l l l l l }
        \textbf{\ac{CNN} variant} & $\bar{T}$ & $\bar{E}$ & $\bar{E}'$ & params & $t_{\text{exec}}$ \\\hline
        SotA - dense & $0.38$ & $1.09$ & $1.02$ & N/A & N/A \\
        Ours ($3\times 3$) - dense  & $\mathbf{0.157}$ & $\mathbf{0.61}$ & $\mathbf{0.54}$ & $100\%$ & \SI{18}{\milli \second}
    \end{tabularx}
    \caption{%
Comparison of the performance of the proposed method and the object detector-based method on the data with high density range.
The best results are highlighted in bold.
}
    \label{tab:densityrange}
    \vspace{-1em}
\end{table}

\subsection{Real-world data generalization}
The dataset described in sec.~\ref{sec:RealData} was used for a qualitative analysis of the proposed method's capability to generalize to real-world data.
The data were split to $30$ images which were used for fine-tuning the trained model and $10$ images that were used for testing.
Fig.~\ref{fig:real2} shows a comparison of the \ac{CNN}'s output on a sample from the real-world data with and without fine-tuning.
After carefully analyzing the results, we conclude that fine-tuning on real-world data significantly improves performance of the \ac{CNN} even with such a relatively small dataset size and that the \ac{CNN} is capable of reproducing the results on the synthetic dataset using this method, which is in line with the results previously reported in~\cite{barisic2022sim2air} for single object detection.

\section{Conclusion}
A novel technique for large-scale visual relative localization of agents within a swarm of \acp{UAV} is introduced in this paper.
We propose a regression-based approach that employs a \ac{CNN} for the estimation of the neighbor density over distance, which enables a more precise distance estimation compared to the state-of-the-art single object detection methods and improves scalability with respect to the number of neighbors.
Our evaluation on both synthetic and real-world data shows the robustness of the proposed approach to various poses of the target \acp{UAV} and its suitability as the primary source of online onboard relative localization for swarm stabilization.
We believe this work will help with the transition from relatively small-scale swarm deployments in controlled environments to large, decentralized, and infrastructure-independent swarms in real-world conditions.
In the future, we aim to design a swarming rule intended to utilize the proposed relative localization method and to extend the real-world dataset to improve the fine-tuning for real-world deployments.
Finally, we intend to carry out real-world swarming experiments using the introduced method in feedback with the swarming rule.


\bibliographystyle{IEEEtran}
\bibliography{main}

\end{document}